\documentclass[10pt, conference, compsocconf]{IEEEtran}
\ifCLASSINFOpdf
\else
\fi
\usepackage{cite}   
\usepackage[utf8]{inputenc} 
\usepackage[T1]{fontenc}    
\usepackage{hyperref}       
\usepackage{url}            
\usepackage{booktabs}       
\usepackage{amsfonts}       
\usepackage{nicefrac}       
\usepackage{microtype}      
\usepackage{todonotes}
\usepackage{pgfplots}
\usepackage{wrapfig}
\usepackage{amsmath,amssymb,amsfonts}
\usepackage{multirow}
\usepackage{bm}             
\newcommand{\ts}{\textsuperscript}
\newcommand\blfootnote[1]{%
  \begingroup
  \renewcommand\thefootnote{}\footnote{#1}%
  \addtocounter{footnote}{-1}%
  \endgroup
}
\usepackage{here}
\usepackage{amsthm}
\DeclareMathOperator{\E}{\mathbb{E}}
\newtheorem{theorem}{Theorem}[section]

\newtheorem{lemma}[theorem]{Lemma}

\theoremstyle{definition}
\usepackage{algorithm}      
\usepackage{algpseudocode}
\let\labelindent\relax
\usepackage{enumitem}
\usepackage{graphicx}
\usepackage{textcomp}
\usepackage{xcolor, soul}
\def\BibTeX{{\rm B\kern-.05em{\sc i\kern-.025em b}\kern-.08em
    T\kern-.1667em\lower.7ex\hbox{E}\kern-.125emX}}

\begin{document}
%
\title{Adversarially Learned Anomaly Detection}




%
\author{\IEEEauthorblockN{
Houssam Zenati\IEEEauthorrefmark{1}\IEEEauthorrefmark{2}\IEEEauthorrefmark{5},
Manon Romain\IEEEauthorrefmark{1}\IEEEauthorrefmark{3}\IEEEauthorrefmark{5},
Chuan-Sheng Foo\IEEEauthorrefmark{1}\IEEEauthorrefmark{5}, 
Bruno Lecouat\IEEEauthorrefmark{4}\IEEEauthorrefmark{5} and
Vijay Chandrasekhar\IEEEauthorrefmark{5}}
\IEEEauthorblockA{\IEEEauthorrefmark{2}CentraleSupélec, houssam.zenati@student.ecp.fr}
\IEEEauthorblockA{\IEEEauthorrefmark{3}\'Ecole Polytechnique, manon.romain@polytechnique.edu}
\IEEEauthorblockA{\IEEEauthorrefmark{4}T\'el\'ecom ParisTech}
\IEEEauthorblockA{\IEEEauthorrefmark{5}Institute for Infocomm Research, A*STAR,
\{foocs,bruno\_lecouat,vijay\}@i2r.a-star.edu.sg}
\IEEEauthorblockA{\IEEEauthorrefmark{1}Authors contributed equally to this work}}



\maketitle

\begin{abstract}
Anomaly detection is a significant and hence well-studied problem. However, developing effective anomaly detection methods for complex and high-dimensional data remains a challenge. As Generative Adversarial Networks (GANs) are able to model the complex high-dimensional distributions of real-world data, they offer a promising approach to address this challenge. In this work, we propose an anomaly detection method, Adversarially Learned Anomaly Detection (ALAD)  based on bi-directional GANs, that derives adversarially learned features for the anomaly detection task. ALAD then uses reconstruction errors based on these adversarially learned features to determine if a data sample is anomalous. 
ALAD builds on recent advances to ensure data-space and latent-space cycle-consistencies and stabilize GAN training, which results in significantly improved anomaly detection performance.
ALAD achieves state-of-the-art performance on a range of image and tabular datasets while being several hundred-fold faster at test time than the only published GAN-based method.\blfootnote{Proceedings of the
20\ts{th} IEEE International Conference on Data Mining, Singapore, 2018. All code and hyperparameters can be found at \url{https://github.com/houssamzenati/Adversarially-Learned-Anomaly-Detection}}

\end{abstract}


%
\IEEEpeerreviewmaketitle

\section{Introduction}

Anomaly detection is a problem of great practical significance across a range of real-world settings, including cyber-security \cite{schubert2014}, manufacturing \cite{marti2015}, fraud detection, and medical imaging \cite{chandola2009}. Fundamentally, anomaly detection methods need to model the patterns in normal data to identify atypical samples. Although anomaly detection is a well-studied problem \cite{chandola2009, zimek2012,noveltydetectionreview2014}, developing effective methods for complex and high-dimensional data remains a challenge.

Generative Adversarial Networks (GANs) \cite{goodfellow2014generative} are a powerful modeling framework for high-dimensional data that could address this challenge. Standard GANs consist of two competing networks, a generator $G$ and discriminator $D$. $G$ models the data by learning a mapping from latent random variables $z$ (drawn from Gaussian or uniform distributions) to the data space, while $D$ learns to distinguish between real data and samples generated by $G$. GANs have been empirically successful as a model for natural images \cite{dcgan2015,overview2017} and are increasingly being used in speech \cite{ganspeechtutorial2018} and medical imaging applications. For example, \cite{anogan2017} proposes a method that uses a standard GAN for anomaly detection on eye images. However, at test time, the method requires solving an optimization problem for each example to find a latent $z$ such that $G(z)$ yields a visually similar image that is also on the image manifold modeled by $G$; this $z$ is then used to compute an anomaly score for the example. The need to solve an optimization problem for every test example makes this method impractical on large datasets or for real-time applications.

In this work, we propose a GAN-based anomaly detection method that is not only effective, but also efficient at test time. 
Specifically, our method utilizes a class of GANs that simultaneously learn an encoder network during training \cite{aligan2016,bigan2016}, thus enabling faster and more efficient inference at test time than \cite{anogan2017}. In addition, we incorporate recent techniques to further improve the encoder network \cite{alice2017} and stabilize GAN training \cite{miyato2018spectral}, and show through ablation studies that these techniques also improve performance on the anomaly detection task. Experiments on a range of high-dimensional tabular and image data demonstrate the efficiency and effectiveness of our approach. 

\section{Related Work}
Anomaly detection has been extensively studied, as surveyed in \cite{chandola2009, zimek2012, noveltydetectionreview2014}. As such, here we give a brief overview and refer the reader to these reviews for a more in-depth discussion. A major class of classic anomaly detection methods are distance-based, using distances to nearest neighbors or clusters in the data to assess if data is anomalous. Such methods rely on having an appropriate distance metric for the data. One-class classification approaches trained only on normal data such as one-class SVMs \cite{ocsvm1999} are also widely used; these methods learn a classification boundary around the normal data. A third class of methods uses fidelity of reconstruction \cite{noveltydetectionreview2014} to determine whether an example is anomalous. Principal component analysis and variants thereof \cite{jolliffe1986, gunter2007, candes2009} are examples of such methods.

More recent works are based on deep neural networks; we note that neural networks have a long history of being used for anomaly detection \cite{nn-noveltydetectionreview2003}. Approaches based on auto-encoders \cite{zhou2017} and variational auto-encoders \cite{an2015} first train a model to reconstruct normal data and subsequently identify anomalies as samples with high reconstruction errors. 
Energy based models \cite{zhai2016} and deep auto-encoding Gaussian mixture models \cite{zong2018deep} have also been explored specifically for the purpose of anomaly detection. Such methods model the data distribution using auto-encoders or similar models, and derive statistical anomaly criterion based on energies or mixtures of Gaussians. 
Finally, GANs have been applied to anomaly detection in the context of medical imaging on retinal images \cite{anogan2017}. However, the methods proposed in \cite{anogan2017} require an inference procedure at test time, where latent variables $z$ are recovered using stochastic gradient descent for every test example. This inference procedure is computationally expensive as every gradient computation requires backpropagation through the generator network.

Our proposed ALAD method is most closely related to the AnoGAN method described in \cite{anogan2017}. However, in contrast to AnoGAN, which uses a standard GAN, ALAD builds upon bi-directional GANs, and hence also includes an encoder network that maps data samples to latent variables. This enables ALAD to avoid the computationally expensive inference procedure required by AnoGAN since the required latent variables can be recovered using a single feed-forward pass through the encoder network at test time. Our anomaly scoring criteria is different from AnoGAN, and we also incorporate recent advances to stabilize the GAN training procedure in ALAD. 

\section{Background}

Standard GANs consist of two neural networks, a \emph{generator} $G$ and \emph{discriminator} $D$, and are trained on a set of $M$ unlabeled data samples $\left\{x^{(i)}\right\}_{i=1}^M$. The generator $G$ maps random variables $z$ sampled from a latent distribution (typically Gaussian or uniform) to the input data space. The discriminator $D$ tries to distinguish real data samples $x^{(i)}$ from samples $G(z)$ generated by $G$. Informally, these two networks $G$ and $D$ are in competition -- $G$ tries to generate samples that resemble real data, while $D$ attempts to discriminate between samples produced by the generator and real data samples. Training a GAN then typically involves taking alternating gradient steps so that $G$ can better ``fool'' $D$, and $D$ can better detect samples from $G$.

Formally, define $p_{\mathcal{X}}(x)$ to be the distribution over data $x$ in the data space $\mathcal{X}$, and $p_{\mathcal{Z}}(z)$ the distribution over latent generator variables $z$ in the latent space $\mathcal{Z}$. Then training a GAN involves finding the discriminator $D$ and the generator $G$ that solve the saddle-point problem \mbox{$\min_{G} \max_{D} V(D, G)$} where
\begin{align*}
V(D, G) &= \E_{x \sim p_{\mathcal{X}}} \big[ \log{D(x)} \big] \\
&+
\E_{z \sim p_{\mathcal{Z}}} \big[ \log{\big(1 - D\left(G(z)\right)\big)} \big].
\label{eq:gan-objective}
\end{align*}
The solutions to this saddle-point problem 
are described by Lemma \ref{lemma:1} (proved in \cite{goodfellow2014generative}), which shows that the optimal generator induces a distribution $p_G(x)$ that matches the true data distribution $p_{\mathcal{X}}(x)$.
\begin{lemma}
\label{lemma:1}
For $G$ fixed, the optimal discriminator $D^{*}_{G}$ is: \[D^{*}_{G} = \frac{p_{\mathcal{X}}(x)}{p_{\mathcal{X}}(x) + p_{G}(x)}.\]
And for this optimal discriminator $D^{*}_{G}$ the global minimum of the virtual training criterion $C(G) = \max_{D} V(D, G)$ is achieved if and only if $p_{G}(x) = p_{\mathcal{X}}(x)$.
\end{lemma}
$D$ and $G$ are typically determined via alternating gradient descent on the parameters of $D$ and $G$, treating the other as fixed, to maximize (for $D$) or minimize (for $G$) $V(D, G)$ accordingly. Having trained the GAN, one can approximately sample from $p_{\mathcal{X}}$ using the generator taking $G(z)$ with $z \sim p_{\mathcal{Z}}$. As explained in Section \ref{secanomalyscore} being able to learn the distribution of the normal data is key for the anomaly detection task. Note that it is not possible to explicitly compute a likelihood or obtain the latent representation for a given data sample $x$ directly using the GAN. 

\section{Adversarially Learned Anomaly Detection}

Given that standard GANs only support efficient sampling, there are several approaches one can take in order to adapt them for anomaly detection. For instance, for a data point $x$, one could use sampling \cite{quantitativeanalysisiclr2017} to estimate the likelihood of $x$ and determine if it is an outlier. However, while sampling from a GAN is efficient, accurate estimation of likelihoods typically requires a large number of samples, thus making the likelihood computation computationally expensive. Another approach is to ``invert'' the generator to find latent variables $z$ that minimize reconstruction error or related objectives by stochastic gradient descent \cite{Zhu2016GenerativeVM, creswell2016, lipton2017, anogan2017}. This approach is also computationally costly as each gradient computation requires backpropagation through the generator network.

\subsection{GAN architecture}

Motivated by computational efficiency, we instead build on a class of GANs that simultaneously learns an encoder network $E$ which maps data samples $x$ to the latent space $z$ during training \cite{bigan2016,aligan2016}. Computing a (approximate) latent representation for a data point $x$ in such models is done simply by passing $x$ through the encoder network. Our models also incorporate recent improvements to improve the encoder network \cite{alice2017} by adding an additional discriminator to encourage cycle-consistency, \emph{i.e.} that $G(E(x)) \approx x$. 

Formally, the BiGAN \cite{bigan2016} and AliGAN \cite{aligan2016} models match the joint distribution $p_{G}(x, z) = p_{\mathcal{Z}}(z) p_{G}(x \vert z)$ and $p_{E}(x, z) = p_{\mathcal{X}}(x) p_{E}(z \vert x)$ with an adversarial discriminator network $D_{xz}$ that takes $x$ and $z$ as inputs. BiGAN and AliGAN determine the discriminator $D_{xz}$, the generator $G$ and the encoder $E$ as the solution to the saddle-point problem: $\min_{G,E} \max_{D_{xz}} V(D_{xz}, E, G)$, with $V(D_{xz}, E, G)$ defined as:
\begin{align*}
V(D_{xz}, E, G) &= \E_{x \sim p_{\mathcal{X}}} \left[ \log{D_{xz}(x,E(x))} \right]  \\&
+
\E_{z \sim p_{\mathcal{Z}}} \left[1 - \log{D_{xz}(G(z),z)} \right]
\end{align*}
The solutions of the saddle-point problem and the distribution matching property $p_{E}(x, z) = p_{G}(x, z)$ are described by Lemma \ref{lemma:2} \cite{aligan2016,bigan2016}:
\begin{lemma}
For $E$ and $G$ fixed, the optimal discriminator $D^{*}_{xz, E, G}$ is: \[D^{*}_{xz} = \frac{p_{E}(x,z)}{p_{E}(x,z) + p_{G}(x,z)}.\]
And for this optimal discriminator $D^{*}_{xz}$ the global minimum of the virtual training criterion $C(E, G) = \max_{D_{xz}} V(D_{xz}, E, G)$ is achieved if and only if $p_{E}(x, z) = p_{G}(x, z)$.
\label{lemma:2}
\end{lemma}
While in theory the joint distributions $p_{E}(x, z)$ and $p_{G}(x, z)$ will be identical, in practice this is often not the case as training does not necessarily converge to a solution of the saddle-point problem. This potentially results in a violation of cycle-consistency so $G(E(x)) \not\approx x$ as highlighted in \cite{alice2017}; such a violation would create issues for reconstruction-based anomaly detection methods.

To solve this problem, the ALICE framework \cite{alice2017} proposes to approximate the conditional entropy $H^{\pi}(x \vert z) = - \E_{\pi (x,z)} \left[ \log{\pi (x \vert z)} \right]$ (where $\pi (x,z)$ is a joint distribution over $x$ and $z$) in an adversarial manner to encourage cycle consistency. This saddle-point problem $\min_{G,E} \max_{D_{xz}} V_{ALICE}(D_{xz}, E, G)$ includes the conditional entropy regularization ($V_{CE}$) on the encoder $E$ and the generator $G$:
\[
V_{ALICE}(D_{xz}, E, G) = V(D_{xz}, E, G) + V_{CE}(E, G)
\]
The conditional entropy regularization imposed on the encoder $E$ and the generator $G$ can be approximated with an additional discriminator network $D_{xx}(x,\hat{x})$
\begin{align*}
V(D_{xx}, E, G) &= \E_{x \sim p_{\mathcal{X}}} \left[ \log{D_{xx}(x,x)} \right] \\&
+ 
\E_{x \sim p_{\mathcal{X}}}  \left[1 - \log{D_{xx}(x,G(E(x)))} \right]
\end{align*}
and \cite{alice2017} formally show that such a discriminator will indeed enforce cycle consistency.


\subsection{Stabilizing GAN training}
\label{sec:stabilize}
To stabilize the training of the baseline ALICE model, we further regularize the conditional distributions by adding another conditional entropy constraint, and apply spectral normalization \cite{miyato2018spectral}.

Formally, we regularize the latent space conditional $H^{\pi}(z \vert x) = - \E_{\pi (x,z)} \left[ \log{\pi (z \vert x)} \right]$ (where $\pi (x,z)$ is a joint distribution over $x$ and $z$) with an additional adversarially learned discriminator $D_{zz}$, with a similar saddle-point objective as follows; the proof of matching of conditionals is a simple adaptation of those presented in \cite{alice2017}. 
\begin{align*}
V(D_{zz}, E, G) &= \E_{z \sim p_{\mathcal{Z}}} \left[ \log{D_{zz}(z,z)} \right] \\&
+ 
\E_{z \sim p_{\mathcal{Z}}}  \left[1 - \log{D_{zz}(z,E(G(z)))} \right]
\end{align*}\label{equation5}

\begin{figure}[htbp]
  \centering
  \includegraphics[width=1\linewidth]{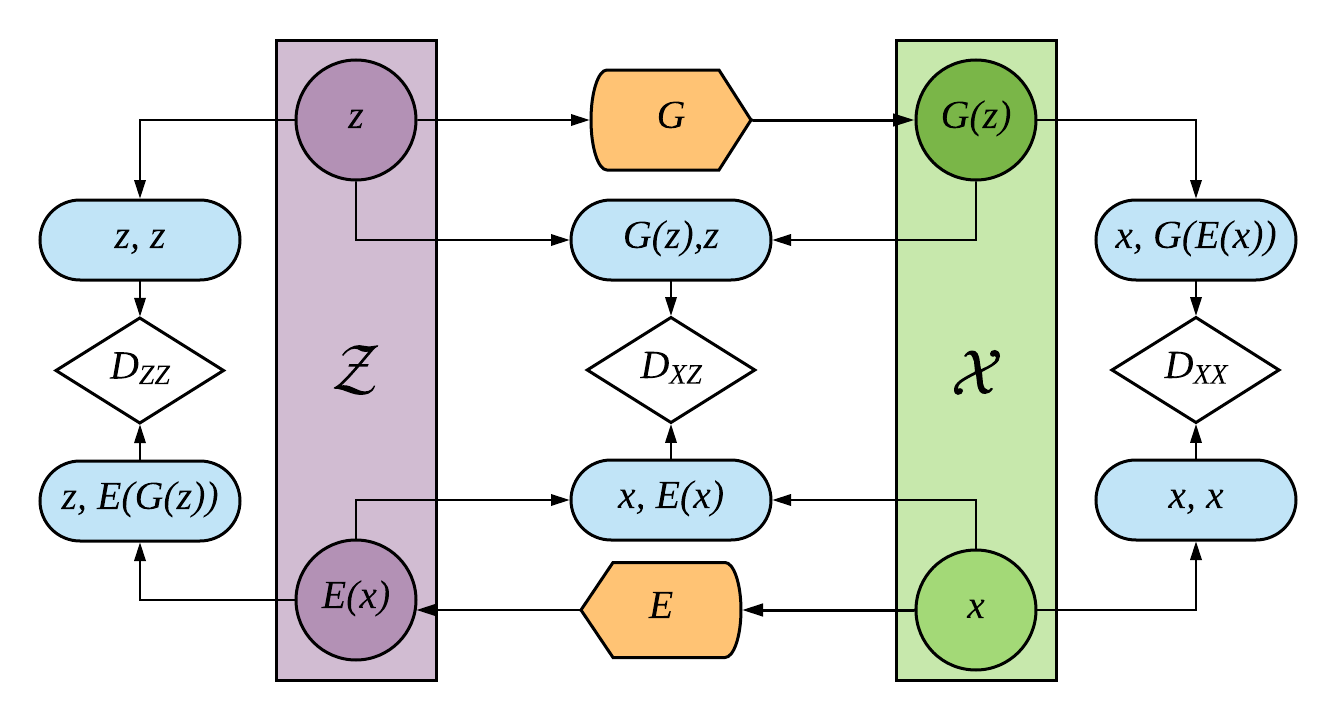}
  \vspace{-8mm}
  \caption{The GAN used in Adversarially Learned Anomaly Detection. $D_{zz}, D_{xz}$ and $D_{xx}$ denote discriminators (white), $G$ the generator (orange), and $E$ the encoder (orange); these networks are simultaneously learned during training.}
\label{fig:gan}
\end{figure}

Putting it all together, our Adversarially Learned Anomaly Detection (ALAD) method solves the following saddle-point problem during training: $\min_{G,E} \max_{D_{xz}, D_{xx}, D_{zz}} V(D_{xz}, D_{xx}, D_{zz}, E, G)$, with $V(D_{xz}, D_{xx}, D_{zz}, E, G)$ defined as 
\begin{multline*}
V(D_{xz}, D_{xx}, D_{zz}, E, G) = \\ 
V(D_{xz}, E, G) + V(D_{xx}, E, G) + V(D_{zz}, E, G).
\end{multline*}
A schematic of this final GAN model is shown in Figure \ref{fig:gan}.



Our addition of spectral normalization is motivated by recent work \cite{Arjovsky2017WassersteinG,improvedwgan2017,miyato2018spectral,selfattentiongan2018} that shows the efficiency of adding Lipschitz constraints on the discriminators of GANs to stabilize the training. In particular \cite{miyato2018spectral} proposes a simple re-parametrization of the weights which turns out to be very effective in practice. They propose to fix the spectral norm of the weight matrix (\emph{i.e.,} its largest eigenvalues) of hidden layers in the discriminator. This method is computationally efficient and stabilizes training. In our experiments, we found that spectral normalization was also beneficial when employed to regularize the encoder (as opposed to the discriminator alone). Note that the original ALICE models \cite{alice2017} did not include this re-parametrization of the weights that we included in our models.

\subsection{Detecting anomalies}
\label{secanomalyscore}

\begin{figure}[htbp]
\centering

\includegraphics[width=0.32\linewidth]{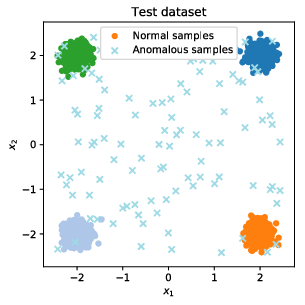}
\includegraphics[width=0.32\linewidth]{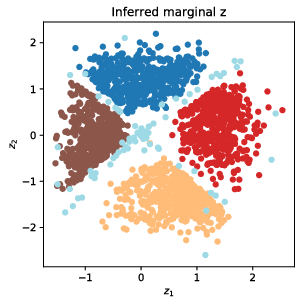}
\includegraphics[width=0.32\linewidth]{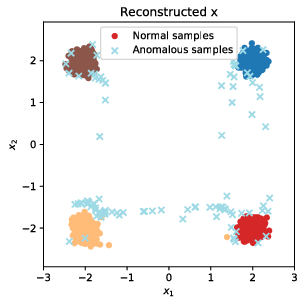}
\vspace{-4mm}
\caption{Toy data example with anomalous samples, encoded representations and reconstructions - circles are normal samples while crosses are anomalous samples.}
\label{fig:toy}
\end{figure}

ALAD is a reconstruction-based anomaly detection technique that evaluates how far a sample is from its reconstruction by the GAN. Normal samples should be accurately reconstructed whereas anomalous samples will likely be poorly reconstructed. This intuition is illustrated in Figure \ref{fig:toy}.

To this end, we first need to effectively model the data distribution: this is achieved using the described GAN, where the generator $G$ is used to learn the distribution of the normal data so that $p_{G}(x) = p_{\mathcal{X}}(x)$, where $p_{G}(x) = \int p_{G}(x \vert z) p_{\mathcal{Z}}(z)dz$. We also need to learn the manifold of the data so as to recover precise latent representations that result in faithful reconstructions of normal samples; the two symmetric conditional entropy cycle-consistency regularization terms in our model help ensure this. 

The other key component of ALAD is an anomaly score that quantifies the distance between the original samples and their reconstructions. The obvious choice of Euclidean distance between the original samples and their reconstructions in data space may not be a reliable measure of dissimilarity. For instance, in the case of images, it can be very noisy because images with similar visual features are not necessarily close to each other in terms of Euclidean distance. 

We instead propose to compute the distance between samples in the feature space of the cycle-consistency discriminator $D_{xx}$, as defined by the layer before the logits; these features are also known as CNN codes. Concretely, having trained a model on the normal data to provide $E$, $G$, $D_{xz}$, $D_{xx}$ and $D_{zz}$, we define a score function $A(x)$ based on the $L_1$ reconstruction error in this feature space:
\[
  A(x) =  \big|\big| f_{xx}(x, x) - f_{xx}(x, G(E(x))) \big|\big|_{1}.
\]
Here, $f(\cdot, \cdot)$ denotes the activations of the layer before the logits (CNN codes) in the $D_{xx}$ network for the given pair of input samples. $A(x)$ captures the discriminators' confidence that a sample is well encoded and reconstructed by our generator and therefore derived from the real data distribution. Samples with larger values of $A(x)$ are deemed more likely to be anomalous. The procedure for computing $A(x)$ is described in Algorithm \ref{alg:1}. Our analyses in Section \ref{section:anomaly-scores} confirm the effectiveness of our proposed anomaly score compared to other possible variants. 

The proposed anomaly score is inspired by the feature-matching loss presented in \cite{salimans2016}. However, unlike \cite{salimans2016} where the feature-matching loss is based on features computed with the discriminator of a standard GAN (that distinguishes between generated samples and real data), the features used by ALAD are computed using the $D_{xx}$ discriminator (that takes pairs of data as input) which does not exist in the standard GAN model considered in \cite{salimans2016}. In addition, we do not use the feature-matching loss within the GAN training procedure, but instead use the concept in a different context of computing an anomaly score at inference time.

\subsection*{Why not use the output of the $D_{xx}$ discriminator?}

We propose that feature loss computed using $D_{xx}$ is preferable over the output of the model $D_{xx}$ for the anomaly score. To see why, consider that $D_{xx}$ is supposed to differentiate between a real sample pair $(x, x)$ and its reconstruction $(x, G(E(x))$. However, at a solution to the GAN saddle-point problem the generator and encoder will perfectly capture the true data and latent variable distribution. In this case, $D_{xx}$ will be unable to discriminate between the real and reconstructed samples, and thus will output a random prediction that would not be an informative anomaly score. Our analyses in Section \ref{section:anomaly-scores} confirm that our proposed anomaly score performs better than the output of $D_{xx}$.

\begin{algorithm}[t]
    \hspace*{\algorithmicindent} \textbf{Input} \quad $\bm{x},  \sim p_{\mathcal{X}_{Test}}(x), E, G, f_{xx}$ where $f_{xx}$ is the feature layer of $D_{xx}$ \\
    \hspace*{\algorithmicindent} \textbf{Output} $A(\bm{x})$, where $A$ is the anomaly score
    \caption{Adversarially Learned Anomaly Detection \label{alg:1}}
    \begin{algorithmic}[1]
    \Procedure{Inference}{}
 	
		\State $\tilde{\bm{z}} \gets E(\bm{x})
			    $ \Comment{Encode samples}
   
     		\State $\hat{\bm{x}} \gets G( \tilde{\bm{z}}),
				$   \Comment{Reconstruct samples}

        \State $f_{\delta} \ \gets f_{xx}(\bm{x}, \hat{\bm{x}})
			$
        \State $f_{\alpha} \ \gets f_{xx}(\bm{x}, \bm{x})
				$

     \State return $\Vert f_{\delta} -  f_{\alpha} \Vert_{1} $ 
                         
\EndProcedure
\end{algorithmic}
\end{algorithm}

\section{Experiments}

\subsection{Experimental setup}

\paragraph{Datasets} We evaluated our Adversarially Learned Anomaly Detection method on publicly available tabular and image datasets (Table \ref{tablestats}). For tabular data, we used the KDDCup99 10\% dataset \cite{Lichman:2013} as well as the Arrhythmia dataset \cite{Lichman:2013}. The KDDCup99 10\% dataset (that we denote by KDD99) is a network intrusion dataset, while the Arrhythmia dataset consists of data in the presence and absence of cardiac arrhythmia as well as classifications into 1 of 16 groups. For the image datasets, we used the SVHN dataset \cite{svhn} containing images of house numbers, and the CIFAR-10 dataset \cite{cifar10} that contains 10 classes of images.

\begin{table}[htbp]
\label{table:1}
\centering
\caption{Statistics of the public benchmark datasets}
\vspace{-2mm}
\label{tablestats}
\begin{tabular}{ccc} \hline
          Dataset & Features & Total Instances  \\ \hline
KDD99      & 121 & 494021                \\
Arrhythmia & 274 & 452          \\ 
SVHN      & 3072 & 99289                 \\
CIFAR-10 & 3072 & 60000            \\ \hline 
\end{tabular}
\end{table}

\paragraph{Data setup and evaluation metrics} For the tabular data, we follow the experimental setups of \cite{zhai2016,zong2018deep}. Due to the high proportion of outliers in the KDD dataset, ``normal'' data are treated as anomalies and the 20\% of samples with the highest anomaly scores $A(x)$ are classified as anomalies (positive class). For the arrhythmia dataset, anomalous classes represent 15\% of the data and therefore the 15\% of samples with the highest anomaly scores are likewise classified as anomalies (positive class). We evaluate our model with the same metrics (Precision, Recall, F1 score) as the state-of-the-art deep learning baselines \cite{zhai2016,zong2018deep}. For the image data, we generate ten different datasets each from SVHN \cite{svhn} and CIFAR-10 \cite{cifar10} by successively treating images from one class as normal and considering images from the remaining 9 as anomalous. For each dataset, we use 80\% of the whole official dataset for training and keep the remaining 20\% as our test set. We further remove 25\% from the training set for a validation set and discard anomalous samples from both training and validation sets (thus setting up a novelty detection task). We evaluate models using the area under the receiver operating curve (AUROC). 
Further details on the experimental setup are provided in the Appendix.  

\subsection{Baselines}

We compare ALAD against several anomaly detection methods, which we briefly describe below.

\begin{description}[style=unboxed,leftmargin=0cm, parsep=10pt]
\item[One Class Support Vector Machines (OC-SVM)] \cite{ocsvm1999} are a classic kernel method for novelty detection that learns a decision boundary around normal examples. We use the radial basis function kernel in our experiments. The $\nu$ parameter is set to the expected anomaly proportion in the dataset, which is assumed to be known, whereas the $\gamma$ parameter is set to $1/m$ where $m$ is the number of input features. 

\item[Isolation Forests (IF)] \cite{iforest2008} are a newer classic machine learning technique that isolates anomalies instead of modeling the distribution of normal data. The method proceeds by first building trees using randomly selected split values across randomly chosen features. Then, the anomaly score is defined to be the average path length from a particular sample to the root. We use default parameters provided by the scikit-learn \cite{scikit-learn} package in our experiments. 

\item[Deep Structured Energy Based Models (DSEBM) \cite{zhai2016}] are a state-of-the-art method that utilize energy-based models. The main idea is to accumulate the energy across layers that are used similarly to those in a denoising autoencoder. In the original paper, two anomaly scoring criteria were investigated: energy and reconstruction error. We include both criteria in our experiments as DSEBM-r (reconstruction) and DSEBM-e (energy). Model details are provided in Appendix \ref{architecture_dsebm}.

\item[Deep Autoencoding Gaussian Mixture Model (DAGMM)] \cite{zong2018deep} is a state-of-the-art autoencoder-based method for anomaly detection. The method first jointly trains an autoencoder to generate a sensible latent space and reconstruction features, as well as an estimator network that will output parameters of a GMM modeling this lower-dimensional latent space. At test time, the likelihood of a sample's latent and reconstruction features as determined using the learned GMM is used as the anomaly detection metric. 

\item[AnoGAN] \cite{anogan2017} is the only published GAN-based method for anomaly detection. The method involves training a DCGAN \cite{dcgan2015}, and at inference time using it to recover a latent representation for each test data sample. The anomaly criterion is a combination of reconstruction and discrimination components. The reconstruction component measures how well the GAN is able to reconstruct the data via the generator, while the discrimination component considers a score based on the discriminator. The original paper \cite{anogan2017} compares two approaches for the anomaly score and we picked the variant which performed best in the paper. For tabular data, we adapted the GAN by using fully connected layers.

\end{description}

\subsection{Experiments on tabular data}

We report in Table \ref{tabular_data_experiments} results on KDD99 \cite{Lichman:2013} as well as Arrhythmia \cite{Lichman:2013} where we see that ALAD is competitive with state-of-the-art methods. Results for OC-SVM, DSEBM, DAGMM were obtained from \cite{zhai2016,zong2018deep} while results for other baselines as well as ALAD are averages over 10 runs.

\begin{table}[htbp]
\caption{Performance on tabular datasets}\label{tabular_data_experiments}
\vspace{-4mm}
\begin{center}
\begin{tabular}{c|c|c|c|c}
{\bf Dataset}  & {\bf Model}  & Precision & Recall & F1 score \\ \hline
\multirow{7}{*}{KDD99}      & OC-SVM     & 0.7457  & 0.8523  & 0.7954\\ 
                            & IF      & 0.9216  & 0.9373  & 0.9294   \\
                            & DSEBM-r    & 0.8521  & 0.6472  & 0.7328\\
                            & DSEBM-e    & 0.8619  & 0.6446  & 0.7399\\
                            & DAGMM      & 0.9297  & 0.9442  & 0.9369\\
                            & AnoGAN     & 0.8786  & 0.8297  & 0.8865\\
                            & {\bf ALAD}      &  \textbf{0.9427}  &   \textbf{0.9577}   &   \textbf{0.9501}    \\ \hline
\multirow{7}{*}{Arrhythmia} & OC-SVM  & \textbf{0.5397}  & 0.4082  & 0.4581  \\ 
                            & IF      & 0.5147	& \textbf{0.5469}	& \textbf{0.5303}  \\
                            & DSEBM-r    & 0.1515  & 0.1513  & 0.1510  \\ 
                            & DSEBM-e   & 0.4667  & 0.4565  & 0.4601  \\ 
                            & DAGMM      & 0.4909  & 0.5078  & 0.4983  \\ 
                            & AnoGAN     &   0.4118 &	0.4375 &	0.4242 \\ 
                            & ALAD      &  0.5000  &   0.5313  &    0.5152 \\
\end{tabular}
\end{center}
\end{table}

Our ALAD method outperforms DAGMM, the best deep learning based method. Interestingly, we observe that Isolation Forests is competitive on the KDD99 dataset and achieves state-of-the-art results on the small Arrhythmia dataset. 
The lack of sufficient training data could have resulted in poorer performance of the data hungry deep learning based methods. On the large KDD99 dataset however, ALAD significantly outperforms all other methods.

\subsection{Experiments on image data}

\begin{figure}[htbp]
\centering
\includegraphics[width=0.35\textwidth]{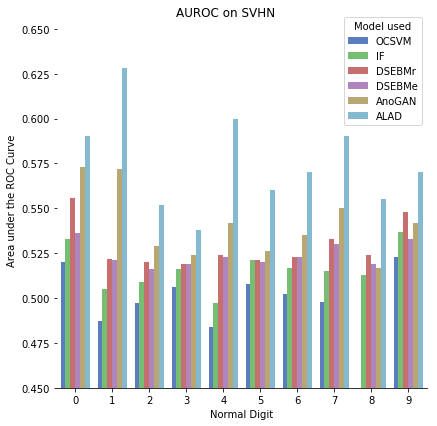}
\vspace{-4mm}
\caption{Area under the ROC curve for the SVHN dataset}
\label{svhn_perf}
\end{figure}

\begin{figure}[htbp]
\centering
\includegraphics[width=0.35\textwidth]{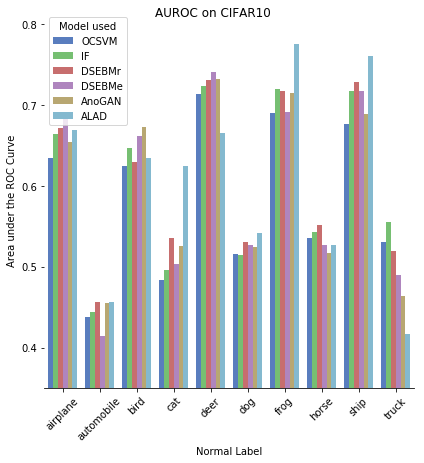}
\vspace{-4mm}
\caption{Area under the ROC curve for the CIFAR-10 dataset}
\label{cifar10_perf}
\end{figure}

We also evaluate our model on SVHN \cite{svhn} as well as CIFAR-10 \cite{Lichman:2013}. We report the results for individual tasks (Figures \ref{svhn_perf} and \ref{cifar10_perf}) and the average performance over all tasks over 3 runs (Table \ref{performances_images}). 

\begin{table}[htbp]
\centering
\caption{Performance on image datasets}
\vspace{-2mm}
\label{performances_images}
\begin{tabular}{c|c|c}
\textbf{Dataset} & \textbf{Model} & \textbf{AUROC}               \\ \hline
SVHN             & OC-SVM         & 0.5027 $\pm$ 0.0132          \\
                 & IF             & 0.5163 $\pm$ 0.0120          \\
                 & DSEBM-r        & 0.5290 $\pm$ 0.0129          \\
                 & DSEBM-e        & 0.5240 $\pm$ 0.0067          \\
                 & AnoGAN         & 0.5410 $\pm$ 0.0193          \\
                 & \textbf{ALAD}  & \textbf{0.5753 $\pm$ 0.0268} \\ \hline
CIFAR-10          & OC-SVM         & 0.5843 $\pm$ 0.0956          \\
                 & IF             & 0.6025 $\pm$ 0.1040          \\
                 & DSEBM-r        & 0.6071 $\pm$ 0.1007          \\
                 & DSEBM-e        & 0.5956 $\pm$ 0.1151          \\
                 & AnoGAN         & 0.5949 $\pm$ 0.1076          \\
                 & \textbf{ALAD}  & \textbf{0.6072 $\pm$ 0.1201}
\end{tabular}
\end{table}

\begin{figure}[htbp]
\centering
\includegraphics[width=0.49\textwidth]{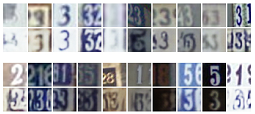}
\vspace{-8mm}
\caption{Error analysis on SVHN when the normal class is the ``3'' digit. 1st row: normal data; 2nd row: reconstruction of normal data; 3rd row: anomalous data; 4th row: reconstruction of anomalous data.}
\label{svhn_reconstructions}
\end{figure}

\begin{figure}[htbp]
\centering
\includegraphics[width=0.49\textwidth]{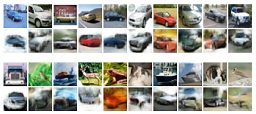}
\vspace{-8mm}
\caption{Error analysis on CIFAR-10 when the normal class is ``Automobile''. 1st row: normal data; 2nd row: reconstruction of normal data; 3rd row: anomalous data; 4th row: reconstruction of anomalous data.}
\label{cifar10_reconstructions}
\end{figure}

We observe that on the SVHN dataset, ALAD outperforms all other methods, while on the CIFAR-10 dataset, ALAD outperforms all other methods on 5 of the 10 tasks, and is competitive on the remaining 5. We also observe that there are some tasks where ALAD does not perform well. On the SVHN dataset, ALAD performs badly on the task where ``3'' is the normal digit. This is possibly because ``3'' is visually similar to several other digits like ``2'' and ``5'' such that the model can produce a relatively good reconstruction even for images from these anomalous digit classes (Figure \ref{svhn_reconstructions}; rows 3, 4), in spite of the fact that these anomalous images tend to be reconstructed to look like the digit ``3'' (that the model was trained on). We observe similar behavior on the CIFAR-10 dataset when we consider the task where ``Automobile'' is the normal class, a task where ALAD performs badly. We see that anomalous samples from the ``Truck'' or ``Ship'' classes are also reconstructed relatively well, even though the reconstructions also look like cars (Figure \ref{cifar10_reconstructions}; rows 3, 4). As a sanity check, we see from the first two rows of Figures \ref{svhn_reconstructions} and \ref{cifar10_reconstructions} that the ALAD model is able to reconstruct examples from the normal class reasonably well.


We also compared the inference time of ALAD and AnoGAN \cite{anogan2017}, the only other GAN-based anomaly detection method (Table \ref{inference_table}). 
On both SVHN \cite{svhn} and CIFAR-10 \cite{cifar10} datasets, the inference time is reported for the first task. We see that ALAD is orders of magnitude faster than AnoGAN.

\begin{table}[htbp]
\centering
\caption{Average inference time (ms) on a GeForce GTX TITAN X}
\vspace{-2mm}
\begin{tabular}{c|c|c|c|c}
{\bf Dataset}     & Batch Size             & AnoGAN & ALAD & Speed Up  \\ \hline
         KDD99    & 50  		   &   1235 & 1.4  &  $\sim$ 900      \\
         Arrhythmia   & 32   	   &   1102 &  41  &  $\sim$ 30   \\
         SVHN     & 32             &   10496 &    10.5     & $\sim$ 1000          \\	
         CIFAR-10   & 32           &   10774     &    10.5     & $\sim$ 1000   
\end{tabular}
\label{inference_table}
\end{table}

\subsection{Ablation studies}

Here we demonstrate the utility of the additional components we used to stabilize the basic ALICE model (Section \ref{sec:stabilize}) by systematically removing each component in turn. Specifically, we repeated our experiments with and without spectral normalization and the additional conditional entropy regularization $H^{\pi}(z \vert x)$.
Our full ALAD model includes both SN (spectral norm) and DL (discriminator in latent space).
\begin{table}[htbp]
\centering
\caption{Ablation Study on KDD99 and Arrhythmia datasets
}
\vspace{-2mm}
\label{tabular_data_ablation}
\begin{tabular}{lccc}
\multicolumn{1}{l|}{\textbf{Model}}     & \multicolumn{1}{c|}{\textbf{Precision}} & \multicolumn{1}{c|}{\textbf{Recall}}   & \textbf{F1 score} \\ \hline
\multicolumn{4}{c}{\textit{KDD99}}                                                                                                      \\ \hline
\multicolumn{1}{l|}{Baseline}           & \multicolumn{1}{c|}{0.948$\pm$0.007}  & \multicolumn{1}{c|}{0.963$\pm$0.007} & 0.955$\pm$0.007 \\
\multicolumn{1}{l|}{Baseline + DL}      & \multicolumn{1}{c|}{0.944$\pm$0.008}  & \multicolumn{1}{c|}{0.959$\pm$0.008} & 0.951$\pm$0.008 \\
\multicolumn{1}{l|}{Baseline + SN} & \multicolumn{1}{c|}{0.942$\pm$0.004}  & \multicolumn{1}{c|}{0.957$\pm$0.004} & 0.949$\pm$0.004 \\ 
\multicolumn{1}{l|}{Baseline + SN + DL}      & \multicolumn{1}{c|}{0.943$\pm$0.002}  & \multicolumn{1}{c|}{0.958$\pm$0.002} & 0.950$\pm$0.002 \\ \hline
\multicolumn{4}{c}{\textit{Arrhythmia}}                                                                                               \\ \hline
\multicolumn{1}{l|}{Baseline}           & \multicolumn{1}{c|}{0.477$\pm$0.039}  & \multicolumn{1}{c|}{0.506$\pm$0.041} & 0.491$\pm$0.040 \\
\multicolumn{1}{l|}{Baseline + DL}      & \multicolumn{1}{c|}{0.497$\pm$0.040}  & \multicolumn{1}{c|}{0.528$\pm$0.043} & 0.512$\pm$0.042 \\
\multicolumn{1}{l|}{Baseline + SN} & \multicolumn{1}{c|}{0.500$\pm$0.021}  & \multicolumn{1}{c|}{0.531$\pm$0.022} & 0.515$\pm$0.021 \\
\multicolumn{1}{l|}{Baseline + SN + DL}      & \multicolumn{1}{c|}{0.482$\pm$0.035}  & \multicolumn{1}{c|}{0.513$\pm$0.037} & 0.497$\pm$0.036 \\

\end{tabular}
\label{ablation_tabular}
\end{table}

The results in Table \ref{ablation_tabular} show that adding the latent penalization improves results on Arrhythmia while it does not affect performance on KDD99. For SVHN and CIFAR-10, we show results averaged over all tasks in Table \ref{ablation_images}; task specific results are shown in the Appendix (Figures \ref{fig:ablation_svhn}, \ref{fig:ablation_cifar10}). On the SVHN tasks, adding the spectral norm (SN) and the discriminator in latent space (DL) improves performance but they had minimal effect on the CIFAR-10 tasks.

\begin{table}[htbp]
\centering
\caption{Ablation Study on SVHN and CIFAR-10 
}
\vspace{-2mm}
\begin{tabular}{lc}
\multicolumn{1}{l|}{\textbf{Model}}     & \textbf{AUROC}      \\ \hline
\multicolumn{2}{c}{\textit{SVHN}}                             \\ \hline
\multicolumn{1}{l|}{Baseline}           & 0.5194 $\pm$ 0.0371 \\
\multicolumn{1}{l|}{Baseline + DL}      & 0.5289 $\pm$ 0.0384 \\
\multicolumn{1}{l|}{Baseline + SN}      & 0.5513 $\pm$ 0.0357 \\
\multicolumn{1}{l|}{Baseline + SN + DL} & 0.5753 $\pm$ 0.0267 \\ \hline
\multicolumn{2}{c}{\textit{CIFAR-10}}                          \\ \hline
\multicolumn{1}{l|}{Baseline}           & 0.5701 $\pm$ 0.1282 \\
\multicolumn{1}{l|}{Baseline + DL}      & 0.5361 $\pm$ 0.1348 \\
\multicolumn{1}{l|}{Baseline + SN}      & 0.5991 $\pm$ 0.1308 \\
\multicolumn{1}{l|}{Baseline + SN + DL} & 0.6072 $\pm$ 0.1201
\end{tabular}
\label{ablation_images}
\end{table}

\subsection{Exploration of different anomaly scores}
\label{section:anomaly-scores}
In this section we evaluate our proposed anomaly score and compare it to other possibilities for reconstruction-based criteria. In particular, we consider the raw output of the discriminator (to which we apply a $log$) referred to as "Logits" but also using non adversarially learned $L_1$ and $L_2$ reconstruction errors. Our approach is denoted by "Features", as it uses features computed by the discriminator. Formally, the scores are defined below, where $x$ is the input sample and $x' = G(E(x))$ is its reconstruction through ALAD:

\begin{itemize}
    \item {\makebox[1.5cm]{$L_1$\hfill}}: $A(x)=||x-x'||_1$
    \item {\makebox[1.5cm]{$L_2$\hfill}}: $A(x)=||x-x'||_2$
    \item {\makebox[1.5cm]{Logits\hfill}}: $A(x)=\log(D_{xx}(x, x'))$
    \item {\makebox[1.5cm]{Features\hfill}}: $A(x)=||f_{xx}(x,x)-f_{xx}(x,x')||_1$
\end{itemize}

\begin{table}[htbp]
\centering
\caption{Different score methods on KDD99 and Arrhythmia datasets}\label{tabular_data_metrics}
\vspace{-2mm}
\begin{tabular}{cccc}
\multicolumn{1}{c|}{\textbf{Model}}    & \multicolumn{1}{c|}{\textbf{Precision}}               & \multicolumn{1}{c|}{\textbf{Recall}}               & \textbf{F1 score}            \\ \hline
\multicolumn{4}{c}{\textit{KDD99}}                                                                                                                                                 \\ \hline
\multicolumn{1}{c|}{$L_1$}              & \multicolumn{1}{c|}{0.9113     $\pm$ 0.0627}          & \multicolumn{1}{c|}{0.9258  $\pm$ 0.0637}          & 0.9185 $\pm$ 0.0632          \\
\multicolumn{1}{c|}{$L_2$}              & \multicolumn{1}{c|}{0.9316     $\pm$ 0.0155}          & \multicolumn{1}{c|}{0.9464  $\pm$ 0.0157}          & 0.9389 $\pm$ 0.0156          \\
\multicolumn{1}{c|}{Logits}            & \multicolumn{1}{c|}{0.9221     $\pm$ 0.0172}          & \multicolumn{1}{c|}{0.9368  $\pm$ 0.0174}          & 0.9294 $\pm$ 0.0173          \\
\multicolumn{1}{c|}{\textbf{Features}} & \multicolumn{1}{c|}{\textbf{0.9427     $\pm$ 0.0018}} & \multicolumn{1}{c|}{\textbf{0.9577  $\pm$ 0.0018}} & \textbf{0.9501 $\pm$ 0.0018} \\ \hline
\multicolumn{4}{c}{\textit{Arrhythmia}}                                                                                                                                            \\ \hline
\multicolumn{1}{c|}{$L_1$}              & \multicolumn{1}{c|}{0.4588     $\pm$ 0.0248}          & \multicolumn{1}{c|}{0.4875  $\pm$ 0.0264}          & 0.4727 $\pm$ 0.0256          \\
\multicolumn{1}{c|}{$L_2$}              & \multicolumn{1}{c|}{0.4529     $\pm$ 0.0206}          & \multicolumn{1}{c|}{0.4813  $\pm$ 0.0219}          & 0.4667 $\pm$ 0.0212          \\
\multicolumn{1}{c|}{Logits}            & \multicolumn{1}{c|}{0.3706     $\pm$ 0.0834}          & \multicolumn{1}{c|}{0.3938  $\pm$ 0.0886}          & 0.3818 $\pm$ 0.0859          \\
\multicolumn{1}{c|}{\textbf{Features}} & \multicolumn{1}{c|}{\textbf{0.5000     $\pm$ 0.0208}} & \multicolumn{1}{c|}{\textbf{0.5313  $\pm$ 0.0221}} & \textbf{0.5152 $\pm$ 0.0214}
\end{tabular}
\end{table}



We observe in Table \ref{tabular_data_metrics} that the adversarially learned features from the reconstruction discriminator are more suited for the anomaly detection task on tabular data.

\begin{table}[htbp]
\centering
\caption{Different score methods on SVHN and CIFAR-10 datasets}\label{image_data_metrics}
\vspace{-2mm}
\begin{tabular}{lc}
\multicolumn{1}{l|}{\textbf{Anomaly Score}} & \textbf{AUROC}      \\ \hline
\multicolumn{2}{c}{\textit{SVHN}}                                 \\ \hline
\multicolumn{1}{l|}{$L_1$}                  & 0.5778 $\pm$ 0.0204 \\
\multicolumn{1}{l|}{$L_2$}                  & 0.5826 $\pm$ 0.0201 \\
\multicolumn{1}{l|}{Logits}                 & 0.5038 $\pm$ 0.0185 \\
\multicolumn{1}{l|}{Features}               & 0.5753 $\pm$ 0.0268 \\ \hline
\multicolumn{2}{c}{\textit{CIFAR-10}}                              \\ \hline
\multicolumn{1}{l|}{$L_1$}                  & 0.6066 $\pm$ 0.1006 \\
\multicolumn{1}{l|}{$L_2$}                  & 0.6012 $\pm$ 0.1088 \\
\multicolumn{1}{l|}{Logits}                 & 0.5396 $\pm$ 0.0783 \\
\multicolumn{1}{l|}{Features}               & 0.6072 $\pm$ 0.1201
\end{tabular}
\end{table}

On image data, the features learned from the reconstruction discriminator lead to better detection of anomalies on the CIFAR-10 dataset, while performing comparably to the $L_1$ and $L_2$ variants on the SVHN dataset (Table \ref{image_data_metrics}); full results are shown in the Appendix (Figures \ref{fig:metrics_svhn}, \ref{fig:metrics_cifar10}).

\section{Conclusion}

We present a GAN-based anomaly detection method ALAD that learns an encoder from the data space to the latent space during training, making it significantly more efficient at test time than the only published GAN method. In addition, we incorporate additional discriminators to improve the encoder, as well as spectral normalization which has been found to stabilize GAN training. Ablation analyses show that these modifications result in improved anomaly detection performance thus confirming their utility. Finally, we showed that ALAD is highly competitive with state-of-the-art anomaly detection methods on a range of tabular and image datasets and often outperforms them. While GANs can be difficult to train, the field is rapidly progressing and our method will directly benefit from any advances that accelerate or stabilize training. 
The effectiveness of ALAD positions GANs as a promising approach for anomaly detection on complex, high-dimensional data; applying ALAD to other data modalities such as speech and sensor data is an interesting direction for future research.

\section*{Acknowledgments}
This research is supported by the Agency for Science, Technology and Research (A*STAR) under its SERC Strategic Funding (A1718g0045).
The computational work for this article was partially performed on resources of the National Supercomputing
Centre, Singapore (\url{https://www.nscc.sg}). We thank the reviewers for their feedback and suggestions, as well as Chunyuan Li and Hao Liu for helpful discussions related to the ALICE method.

\bibliography{icdm_2018}{}
\bibliographystyle{IEEEtran}

\appendix

\subsection{CIFAR-10 and SVHN Experimental Details}
\textbf{Preprocessing:} Pixels were scaled to be in range [-1,1]. 

\subsubsection{DSEBM}
\label{architecture_dsebm}
For both CIFAR-10 and SVHN, we used the architecture suggested in \cite{zong2018deep}: one convolutional layer with kernel of size 3, strides of size 2, 64 filters and "same" padding, one max-pooling layer and one fully connected layer with 128 units. \\

\subsubsection{AnoGAN}
We took the official DCGAN architecture and hyper-parameters for these experiments. For the anomaly detection task, we took the same hyperparameters as the original paper. Similarly to ALAD, we used exponential moving average for inference with a decay of 0.999.  \\

\subsubsection{ALAD}
The outputs of the underlined layer in the discriminator were used for the anomaly score. All convolutional layers have "same" padding unless specified otherwise.

\begin{table}[htbp]
\tiny
\centering
\begin{tabular}{lllllll}
\hline
Operation                                     & Kernel       & Strides      & \begin{tabular}[c]{@{}l@{}}Filters\\ Units\end{tabular} & \begin{tabular}[c]{@{}l@{}}Non\\ Linearity\end{tabular}       & \begin{tabular}[c]{@{}l@{}}Batch\\ Norm.\end{tabular}  \\ \hline
\multicolumn{7}{l}{$\bm{E(z)}$} \\
Conv2D                                   & $4\times4$ & $2\times2$ & 128                   & LReLU  & $\surd$           \\
Conv2D                                   & $4\times4$ & $2\times2$ & 256                  & LReLU  & $\surd$           \\
Conv2D                                   & $4\times4$ & $2\times2$ & 512                   & LReLU  & $\surd$       \\
Conv2D$^\star$                                    & $4\times4$ & $1\times1$ & 100                    & None  & $\times$      \\
\multicolumn{7}{l}{$\bm{G(z)}$ } \\
Trans. Conv2D$^\star$                                  & $4\times4$ & $2\times2$ & 512                   & ReLU   & $\surd$      \\
Trans. Conv2D                                   & $4\times4$ & $2\times2$ & 256                 & ReLU    & $\surd$       \\
Trans. Conv2D                                   & $4\times4$ & $2\times2$ & 128                    & ReLU     & $\surd$   \\
Trans. Conv2D                                   & $4\times4$ & $2\times2$ & 3                    & Tanh    & $\surd$    \\
\multicolumn{7}{l}{$\bm{D_{xz}(x, z)}$}  \\
\textit{Only on $x$}&&&&&&\\
Conv2D                 & $4\times4$ & $2\times2$ & 128                   & LReLU  & $\times$  \\
Conv2D                 & $4\times4$ & $2\times2$ & 256                    & LReLU & $\surd$  \\
Conv2D                 & $4\times4$ & $2\times2$ & 512                    & LReLU & $\surd$  \\
\textit{Only on $z$}&&&&&&\\
Conv2D$^\dagger$                 & $4\times4$ & $2\times2$ & 512                    & LReLU & $\times$  \\
Conv2D$^\dagger$                 & $4\times4$ & $2\times2$ & 512                  & LReLU  & $\times$   \\
\multicolumn{7}{l}{\textit{Concatenate outputs}} \\
Conv2D$^\dagger$                 & $1\times1$ & $1\times1$ & 1024                   & LReLU  & $\times$  \\
Conv2D                          & $1\times1$ & $1\times1$ & 1                       & LReLU & $\times$  \\
\multicolumn{7}{l}{$\bm{D_{xx}(x, x')}$ } \\
\multicolumn{7}{l}{\textit{Concatenate x and x'}} \\
Conv2D$^\dagger$                  & $5\times5$ & $2\times2$ & 64                   & LReLU  & $\times$  \\
\underline{Conv2D$^\dagger$}                & $5\times5$ & $2\times2$ & 128                    & LReLU & $\times$  \\
Dense                               &               &             & 1                   & None  & $\times$  \\
\multicolumn{7}{l}{$\bm{D_{zz}(z, z')}$}\\
\multicolumn{7}{l}{\textit{Concatenate z and z'}} \\
Dense$^\dagger$              &              &              & 64                    & LReLU  & $\times$   \\
Dense$^\dagger$             &              &              & 32                     & LReLU  & $\times$ \\
Dense$^\dagger$             &              &              & 1                      & LReLU  & $\times$ \\ \hline
Optimizer                   & \multicolumn{5}{l}{Adam($\alpha=2*10^{-4}$, $\beta_{1}=0.5$)}                                                   \\
Batch size                  & \multicolumn{5}{l}{32}                                                      \\
Latent dimension          	& \multicolumn{5}{l}{100}                                                      \\
Max Epochs       			& \multicolumn{5}{l}{100}                                                      \\
Patience         			& \multicolumn{5}{l}{10}                                                      \\
LReLU slope            & \multicolumn{5}{l}{0.2}                                                      \\
Weight \& & \multicolumn{5}{l}{Isotropic gaussian ($\mu=0$, $\sigma=0.01$)} \\
bias initialization & \multicolumn{5}{l}{Constant(0)}\\ \hline                         
\end{tabular}
\caption{CIFAR-10 and SVHN ALAD architecture and hyperparameters 
($^\dagger$ dropout, $^\star$ valid padding)}\label{aladimages}
\end{table}

\subsection{KDD99 Experiment Details}

\textbf{Preprocessing:} The dataset contains samples of 41 dimensions, where 34 of them are continuous and 7 are categorical. For categorical features, we further used one-hot representation to encode them; we obtained a total of 121 features after this encoding. 
\\

\begin{table}[htbp]
\tiny
\centering
\begin{tabular}{llll}
\hline
Operation                   & Units        & \begin{tabular}[c]{@{}l@{}}Non\\ Linearity\end{tabular}        & Dropout       \\ \hline
\multicolumn{4}{l}{$\bm{G(z)}$} \\
Dense                       & 64          & ReLU                 & 0.0           \\
Dense                       & 128          & ReLU                 & 0.0           \\
Dense                       & 121          & None               & 0.0           \\
\multicolumn{4}{l}{$\bm{D(x)}$} \\
Dense                       & 256          & LReLU           & 0.2          \\
Dense                       & 128        & LReLU             & 0.2          \\
Dense                       & 128        & LReLU             & 0.2          \\
Dense                       & 1            & Sigmoid              & 0.0           \\ \hline
Optimizer                   & \multicolumn{3}{l}{Adam($\alpha=10^{-5}$, $\beta_{1}=0.5$)}                                                   \\
Batch size                  & \multicolumn{3}{l}{50}                                                      \\
Latent dimension          	& \multicolumn{3}{l}{32}                                                      \\
Epochs          			& \multicolumn{3}{l}{100}                                                      \\
LReLU slope            & \multicolumn{3}{l}{0.2}                                                      \\
Weight, bias initialization & \multicolumn{3}{l}{Xavier Initializer, Constant(0)}     \\ \hline 
\end{tabular}
\caption{KDD99 GAN Architecture and hyperparameters}\label{kddgan}
\end{table}

\begin{table}[htbp]
\tiny
\centering
\begin{tabular}{lllll}
\hline
Operation                                     & Units        & \begin{tabular}[c]{@{}l@{}}Non\\ Linearity\end{tabular}         & \begin{tabular}[c]{@{}l@{}}Batch\\ Norm.\end{tabular}  &Dropout       \\ \hline
\multicolumn{4}{l}{$\bm{E(x)}$} \\
Dense                                         & 64          & LReLU               & $\times$  &    0.0     \\
Dense                                         & 1          & None               & $\times$       &  0.0 \\
\multicolumn{4}{l}{$\bm{G(z)}$} \\
Dense                       & 64          & ReLU                 & $\times$       &   0.0 \\
Dense                       & 128          & ReLU                & $\times$      &    0.0 \\
Dense                       & 121          & None               &$\times$       & 0.0  \\
\multicolumn{4}{l}{$\bm{D_{xz}(x, z)}$}\\
\textit{Only on x}                                          &              &                      &       &        \\
Dense                       & 128          & LReLU          & $\surd$       &  0.0  \\
\textit{Only on z}                                          &              &                      &           &    \\
Dense                       & 128          & LReLU          &$\times$     &  0.5    \\
\multicolumn{4}{l}{ \textit{Concatenate outputs}} \\
Dense                       & 128          & LReLU          & $\times$       &  0.5  \\
Dense                       & 1          & Sigmoid          & $\times$       &  0.0  \\
\multicolumn{4}{l}{$\bm{D_{xx}(x, x')}$}\\
\multicolumn{4}{l}{\textit{Concatenate x and x'}} \\
\underline{Dense}                       & 128          & LReLU          & $\times$     &   0.2   \\
Dense                       & 1          & Sigmoid          & $\times$          &  0.0\\
\multicolumn{4}{l}{$\bm{D_{zz}(z, z')}$}\\
\multicolumn{4}{l}{\textit{Concatenate z and z'}} \\
Dense                       & 32          & LReLU          &$\times$    &    0.2\\
Dense                       & 1          & Sigmoid          & $\times$     &     0.0 \\ \hline

Optimizer                   & \multicolumn{4}{l}{Adam($\alpha=10^{-5}$, $\beta_{1}=0.5$)}                                                   \\
Batch size                  & \multicolumn{4}{l}{50}                                                      \\
Latent dimension          	& \multicolumn{4}{l}{32}                                                      \\
Epochs          			& \multicolumn{4}{l}{100}                                                      \\
LReLU slope            & \multicolumn{4}{l}{0.2}                                                      \\
Weight, bias init. & \multicolumn{4}{l}{Xavier Initializer, Constant(0)}     \\ \hline 
\end{tabular}
\caption{KDD99 ALAD Architecture and hyperparameters}\label{kddbigan}
\end{table}

\subsection{Arrhythmia Experiment Details}

\textbf{Preprocessing:} The dataset contains samples of 274 dimensions. We applied our methods on raw data.
\\

\begin{table}[htbp]
\tiny
\centering
\begin{tabular}{llll}
\hline
Operation   & Units        & \begin{tabular}[c]{@{}l@{}}Non\\ Linearity\end{tabular}        & Dropout       \\ \hline
\multicolumn{4}{l}{$\bm{G(z)}$} \\
Dense                       & 128          & ReLU                 & 0.0           \\
Dense                       & 256          & ReLU                 & 0.0           \\
Dense                       & 274          & None               & 0.0           \\
\multicolumn{4}{l}{$\bm{D(x)}$} \\
Dense                       & 256          & LReLU           & 0.2          \\
Dense                       & 128        & LReLU             & 0.5          \\
Dense                       & 1            & Sigmoid              & 0.0           \\ \hline
Optimizer                   & \multicolumn{3}{l}{Adam($\alpha=10^{-5}$, $\beta_{1}=0.5$)}                                                   \\
Batch size                  & \multicolumn{3}{l}{32}                                                      \\
Latent dimension          	& \multicolumn{3}{l}{64}                                                      \\
Epochs          			& \multicolumn{3}{l}{1000}                                                      \\
LReLU slope            & \multicolumn{3}{l}{0.2}                                                      \\
Weight, bias initialization & \multicolumn{3}{l}{Xavier Initializer, Constant(0)}     \\ \hline 
\end{tabular}
\caption{Arrhythmia GAN Architecture and hyperparameters}\label{arrhythmiagan}
\end{table}

\begin{table}[htbp]
\tiny
\centering
\begin{tabular}{lllll}
\hline
Operation                                     & Units        & \begin{tabular}[c]{@{}l@{}}Non\\ Linearity\end{tabular}        & \begin{tabular}[c]{@{}l@{}}Batch\\ Norm.\end{tabular} &Dropout       \\ \hline
\multicolumn{4}{l}{$\bm{E(x)}$} \\
Dense                                         & 256          & LReLU               & $\times$  &   0.0      \\
Dense                                         & 128          & LReLU               & $\times$  &    0.0     \\
Dense                                         & 64          & None               & $\times$       & 0.0  \\
\multicolumn{4}{l}{$\bm{G(z)}$} \\
Dense                       & 128          & ReLU                 & $\times$       &  0.0  \\
Dense                       & 256          & ReLU                & $\times$      &  0.0   \\
Dense                       & 274          & None               &$\times$       &  0.0 \\
\multicolumn{4}{l}{$\bm{D_{xz}(x, z)}$}\\
\textit{Only x}                                          &              &                      &       &        \\
Dense                       & 128          & LReLU          & $\surd$       &  0.0  \\
\textit{Only z}                              &              &                      &       &        \\
Dense                       & 128          & LReLU          &$\times$     &  0.5    \\
\multicolumn{4}{l}{ \textit{Concatenate outputs}} \\

Dense                       & 256          & LReLU          & $\times$       &  0.5  \\
Dense                       & 1          & Sigmoid          & $\times$       &  0.0  \\
\multicolumn{4}{l}{$\bm{D_{xx}(x, x')}$}\\
\multicolumn{4}{l}{\textit{Concatenate x and x'}} \\
Dense                       & 256          & LReLU          & $\times$     &    0.2  \\
\underline{Dense}                       & 128          & LReLU          & $\times$     &   0.2   \\
Dense                       & 1          & Sigmoid          & $\times$          & 0.0 \\
\multicolumn{4}{l}{$\bm{D_{zz}(z, z')}$}\\
\multicolumn{4}{l}{\textit{Concatenate z and z'}} \\
Dense                       & 64          & LReLU          & $\times$    & 0.2     \\
Dense                       & 32          & LReLU          &$\times$    &    0.2\\
Dense                       & 1          & Sigmoid          & $\times$     &  0.0    \\ \hline

Optimizer                   & \multicolumn{4}{l}{Adam($\alpha=10^{-5}$, $\beta_{1}=0.5$)}                                                   \\
Batch size                  & \multicolumn{4}{l}{32}                                                      \\
Latent dimension          	& \multicolumn{4}{l}{64}                                                      \\
Epochs          			& \multicolumn{4}{l}{1000}                                                      \\
LReLU slope            & \multicolumn{4}{l}{0.2}                                                      \\
Weight, bias init. & \multicolumn{4}{l}{Xavier Initializer, Constant(0)}     \\ \hline 
\end{tabular}
\caption{Arrhythmia ALAD Architecture and hyperparameters}\label{arrhythmiabigan}
\end{table}

\begin{figure}[htbp]
\centering
\includegraphics[width=0.28\textwidth]{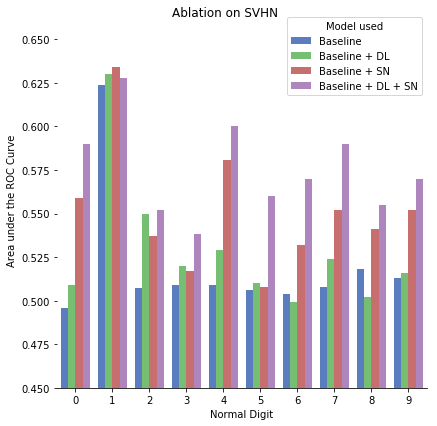}
\vspace{-4mm}
\caption{Ablation study on SVHN}
\label{fig:ablation_svhn}
\end{figure}

\begin{figure}[htbp]
\centering
\includegraphics[width=0.28\textwidth]{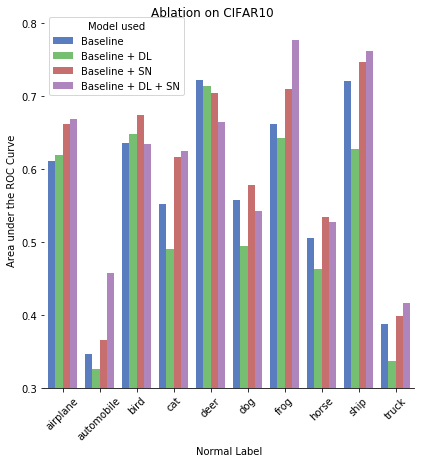}
\vspace{-4mm}
\caption{Ablation study on CIFAR-10}
\label{fig:ablation_cifar10}
\end{figure}

\begin{figure}[htbp]
\centering
\includegraphics[width=0.28\textwidth]{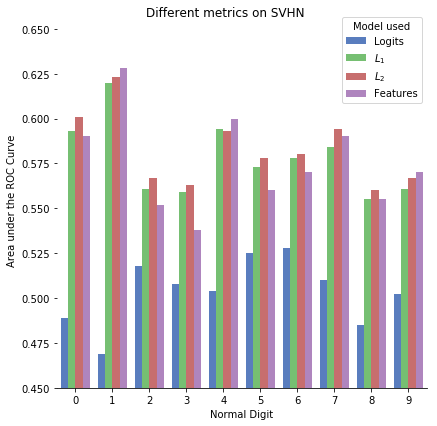}
\vspace{-4mm}
\caption{Performance of different anomaly scores on SVHN}
\label{fig:metrics_svhn}
\end{figure}

\begin{figure}[htbp]
\centering
\includegraphics[width=0.28\textwidth]{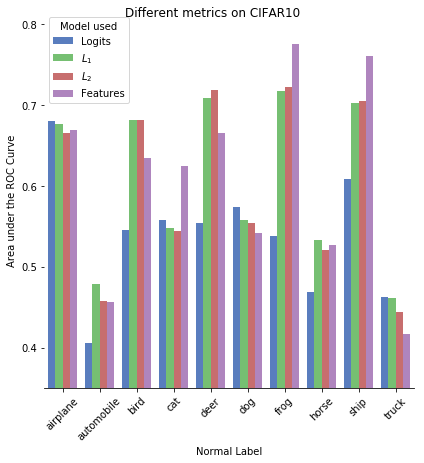}
\vspace{-4mm}
\caption{Performance of different anomaly scores on CIFAR-10}
\label{fig:metrics_cifar10}
\end{figure}

\end{document}